\documentclass[conference]{IEEEtran}

\usepackage[T1]{fontenc}

\usepackage{amsmath,amssymb,amsfonts}
\usepackage{csquotes, xpatch}
\usepackage{microtype}

\usepackage[noadjust]{cite}
\usepackage{graphicx}
\graphicspath{ {figs} {}}

\title{Akkumula: Evidence accumulation driver models with Spiking Neural Networks}

\author{\IEEEauthorblockN{Alberto Morando}
\IEEEauthorblockA{
    Autoliv Development AB\\
    Email: alberto.morando@autoliv.com}
    }

\begin{document}
\maketitle

\begin{abstract}
Processes of evidence accumulation can make driver models more realistic, by explaining how drivers adjust their actions based on perceptual inputs and decision boundaries. The absence of a standard modelling approach limits their adoption; existing methods are hand-crafted, hard to adapt, and computationally inefficient. This paper presents Akkumula, an evidence accumulation modelling framework that uses Spiking Neural Networks and other deep learning techniques. Tested on data from a test-track experiment, the model can reproduce the time course of braking, accelerating, and steering. Akkumula integrates with existing machine learning architectures, scales to large datasets, adapts to different driving scenarios, and keeps its internal logic relatively transparent.
\end{abstract}

\section{Introduction}
Driving, like many other daily activities, is a continuous process of evidence accumulation and a series of intermittent control \cite{markkulaSustainedSensorimotorControl2018}. Inspired by the work of Markkula et al. \cite{markkulaSustainedSensorimotorControl2018} and Boda et al. \cite{bodaComputationalDriverModel2020}, I present here the computational library \emph{Akkumula} to build evidence accumulation models in a modern way, borrowing concepts from cognitive neuroscience and techniques from deep learning.

In \emph{cognitive science}, evidence accumulation models are known as Drift Diffusion Models (DDMs). They are mathematical abstractions of the decision process, often applied to well-constrained lab experiments to study decision speed and accuracy (e.g., single-choice tasks paradigm) \cite{ratcliffDiffusionDecisionModel2016}. At their core, DDMs describe decision-making as a process where noisy sensory input is accumulated over time until reaching a decision boundary. When multiple choices (e.g., two-choice paradigm) are available, evidence builds up for different options until there is enough support for one of them \cite{ratcliffDiffusionDecisionModel2008}. DDMs, in their traditional form or more advanced definitions (e.g., variable drift), have been applied to traffic safety to model decisions to brake \cite{svardQuantitativeDriverModel2017,ratcliffModelingSimpleDriving2014,bianchipiccininiHowDriversRespond2019}, to steer \cite{markkulaSustainedSensorimotorControl2018}, cross the road \cite{gilesZebraCrossingModelling2024,pekkanenVariableDriftDiffusionModels2022}, and negotiate intersections \cite{bodaComputationalDriverModel2020,zgonnikovShouldStayShould2022}. While most studies have focused on modeling one decision point (e.g., brake reaction time), few have modeled the timeseries of brake adjustments \cite{markkulaSustainedSensorimotorControl2018,bodaComputationalDriverModel2020}.

In \emph{neuroscience}, neuronal dynamics models, such as the leaky integrate-and-fire (LIF) neuron \cite{NeuronalDynamicsNeuroscience}, are mathematical abstractions of the temporal flow of electrical information between neurons. LIF neurons accumulate incoming electrical inputs over time from neighboring neurons and generate an output electrical signal (spike) only if the internal potential exceeds a threshold. The bounded integration of inputs makes LIF neurons and DDMs similar in their operation. Neuronal dynamics alone is not used for driver behavior modeling, because the interest often lies in understanding complex cognitive processes rather than focusing on low-level biological activities.

In \emph{deep learning}, concepts from neuronal dynamics can offer a new approach for modelling cognitive processing as an alternative to DDMs. Spiking Neural Networks (SNNs) are a type of Artificial Neural Network (ANN) with close resemblance to the biological brain \cite{eshraghianTrainingSpikingNeural2023}. The artificial neurons in SNNs implement functions derived from neuronal dynamics (e.g., the LIF neuron). While ANNs exchange real-valued signals between neurons, SNNs transmit information through discrete spikes, mirroring the brain's communication strategy. While ANNs excel in predictions in many domains, SNNs are considered more biological plausible but they are still in the early stages of research and development. Luckily, many techniques from advancements in deep learning already apply to SNNs \cite{eshraghianTrainingSpikingNeural2023}, including compatibility with widespread programming libraries (and most ANN modules), fast computation of large batches of data, and efficient multi-parameter gradient-based optimization (thanks to the introduction of surrogate gradients \cite{neftciSurrogateGradientLearning2019}). This approach brings several advantages compared to currently available DDMs applications, which rely on hand-crafted code that is not easily adaptable to different use cases, does not scale well with large data, and relies on computationally inefficient gradient-free optimization.

In this paper, I combine cognitive and neuroscience principles of threshold-based signal integration with deep learning techniques to create a biologically plausible and computationally powerful framework for driver modeling. Akkumula  connects theoretical decision-making concepts with traffic safety applications, expanding the ability to model multiple aspects of driving behavior in dynamic scenarios.

\section{Methods}
\subsection{Model}
\begin{figure}[tb]
    \includegraphics[width=\columnwidth, keepaspectratio]{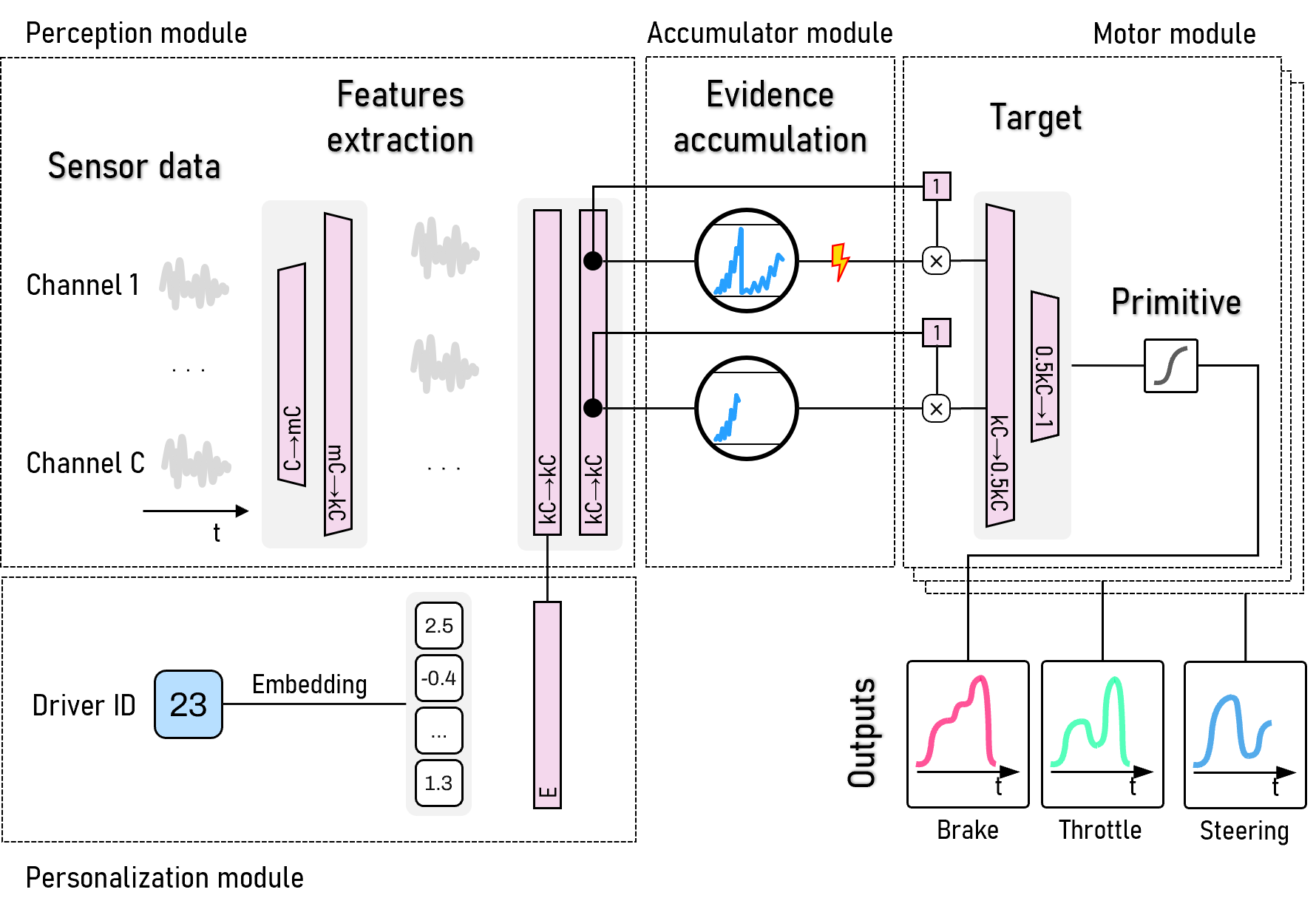}
        \centering
        \caption{Model structure and depiction of the main modules. Emphasis is put on overall structure rather than on details of single components. There are three copies of the motor module, each dedicated to a specific control action (brake, throttle, and steering).}
        \label{fig-model-structure}
\end{figure}

The driver model is built using modules included in the Akkumula framework: the perception, the accumulator, and the motor module (Fig.~\ref{fig-model-structure}). The personalization module is optional, but it allows to learn individual driver behavior traits from the complete data pool. Each module is itself a composition of functions or submodules—general ANN functions from \emph{PyTorch} (v. 2.7) \cite{anselPyTorch2Faster2024} (e.g., linear layers, non-linear activations) and specific SNN functions from \emph{SpikingJelly} (v. 0.0.0.0.14) \cite{fangSpikingJellyOpensourceMachine2023,fangwei123456Fangwei123456Spikingjelly2024} (e.g., LIF neurons).

The model processes the input timeseries sequentially (one timestamp at a time), and at each timestamp, it outputs the predicted motor control for brake, throttle, and steering. In this respect, the model resembles a Recurrent Neural Networks (RNNs; a class of ANNs for sequential data). Unlike RNNs, the model does not have designated memory (hidden) cells to retain information from prior inputs to influence the current input and output. Instead, the model stores prior information in the form of accumulated evidence in the accumulator module.

\subsubsection{Perception module}
The perception module processes information from the vehicle and environment to extract a set of (non-linear) features that are passed along the model at each timestamp (Fig.~\ref{fig-model-structure}). At this stage, the module consists of two linear, fully-connected layers (with bias) with the \emph{mish} activation \cite{misraMishSelfRegularized2020} in between and a \emph{tanh} activation at the output. The mish function was chosen to prevent dead neurons (a situation where some neurons remain inactive during training) by maintaining a small, non-zero gradient at negative inputs. The tanh was chosen to emulate sensor saturation (as in \cite{pekkanenVariableDriftDiffusionModels2022}), while preventing dead neurons. The first linear layer in the module expands the $C$ input features to $mC$ ($m=7$). Then, the second linear layer compresses the features to $kC$ ($k=5$; Fig.~\ref{fig-model-structure}). A two-layer network is shallow by deep learning standards, but it performed well in practice with the training data used.

Previous work on evidence accumulation driver models has used a limited set of predefined metrics for locomotion and obstacle avoidance (e.g., invariants \cite{markkulaSustainedSensorimotorControl2018} such as longitudinal looming \cite{leeTheoryVisualControl1976} and lateral shift \cite{fajenGuidingLocomotionComplex2013}). However, as the driving scenarios change and become more complex, hand-crafted features can be insufficient to explain driver's decision and control over time (e.g., looming can explain brake behavior in conflict situations \cite{svardQuantitativeDriverModel2017,markkulaFarewellBrakeReaction2016}, but alone it may not explain regular, yet complex, traffic negotiations with other road users).

The intent with the perception module is that the network would extract relevant features automatically, presumably "discovering" some of those invariants that have been documented in the literature. Although the exact meaning of these generated features may be nebulous, they are likely more comprehensive and context-dependent due to the interconnected inputs and intermediary representations in the fully-connected layers. Hand-crafted features could still be added to automatically generated ones, or the perception module could be skipped altogether, if preferred.

All linear layers' parameters are learned during training. The number of layers, and the feature multipliers $m$ and $k$ are hyperparameters that can be tuned.

\subsubsection{Personalization module}
The personalization module defines a way to account for specific drivers' traits. In classical statistical modeling, it is best practice to account for individual subject's variation with multi-level (hierarchical or random effects) models \cite{kruschkeBayesianEstimationHierarchical2015}. There seems to be no equivalent and established way of integrating such information in neural networks. One possible alternative is the use of embeddings, which encode categorical features as continuous vectors \cite{mikolovEfficientEstimationWord2013}.

Embeddings are essentially look-up tables, in the sense that each row of size $E$ (here $E=10$) corresponds to one of the encoded categories (the driver's identification code). The relative distance between learned vectors in the embedding space has a potential meaningful relationship; for example, vectors close together could signify people with similar driving styles. Also, each component of the vector could be interpreted as a marker that defines a specific style (e.g., sensation seeking \cite{sagbergReviewResearchDriving2015}). One way to inspect the structure of the embeddings is to project it to a lower dimension space using Principal Component Analysis (PCA).

The embeddings are learned during training. The personalization module is integrated at the sensory level (Fig.~\ref{fig-model-structure}), with the idea that different people would seek, interpret, and accumulate evidence in their individual way. The drivers' embeddings are concatenated with the $kC$ features from the perception module, resulting in an array of $kC+E$ elements. Next, the $kC+E$ elements are compressed to $kC$ with a single linear layer, effectively incorporating driver's specific sensory characteristics to the extracted input features.

During validation or testing, the average of the embeddings will be assigned to the unknown (out of training data) driver. This approach is close to classic multi-level models, where the central tendency of the population is close to the average of the specific subjects' variation.

\subsection{Accumulator module}


The accumulator module is the key part of the model. It consists of multiple customized LIF neurons running in parallel, one for each of the $kC$ features from the perception module (Fig.~\ref{fig-model-structure}). The charge equation of the LIF neurons \cite[Ch.~5]{NeuronalDynamicsNeuroscience} was modified to align with the DDM's formulation \cite{ratcliffDiffusionDecisionModel2016}; DDMs have been used before in traffic research \cite{markkulaSustainedSensorimotorControl2018,bodaComputationalDriverModel2020,svardQuantitativeDriverModel2017} and they may be more familiar to traffic researchers. The charge mechanism in the custom LIF neuron is governed by the equations:

\begin{align}
        \frac{dv}{dt} &= -\rho v + y \label{eq-accumulator-1}\\
        y &= f(x) = a + bx \label{eq-accumulator-2}
\end{align}

where $v$ is the membrane potential (or the activity \cite{markkulaSustainedSensorimotorControl2018}); $\rho$ is the leakage (a proportion of the voltage that is lost at each timestamp); $y$ is the input to the neuron. The input $y$ is computed from the sensory inputs $x$ as a linear combination of a bias $a$ and a gain $b$. The bias can shift the accumulation trajectory toward (or away) the firing threshold, and the gain is the rate of accumulation (drift rate in \cite{ratcliffDiffusionDecisionModel2016}). The function $f(x)$ is not a linear layer, as defined in deep learning frameworks, because the $kC$ features remain unmixed. The equation (\ref{eq-accumulator-1}) is solved at each time step in a discretized form with the Euler step:

\begin{equation}
    v_{t+1} = v_t + [-\rho v_t + f(x)] \Delta t
    \label{eq-euler-step}
\end{equation}

When the neuron potential exceeds a specified threshold (set to 1, as it is common practice), the neuron spikes. After spiking, its voltage is reset to a set value, $v_\text{reset}$, which can be 0, or any positive or negative value. The output of the accumulator module is binary \{0: dormant, 1: spike\}. The parameters $\rho$, $a$, $b$, and $v_\text{reset}$ are learned during training.





\subsection{Motor module}

The motor module implements intermittent motor control (Fig.~\ref{fig-model-structure}). The superposition of smooth, elementary motor primitives can generate complex motor trajectories and it is neurobiologically plausible \cite{markkulaSustainedSensorimotorControl2018,bodaComputationalDriverModel2020,degallierrochatDiscreteRhythmicMotor2010,giszterMotorPrimitivesNew2015}. There are several motor primitives choices, all typically resembling a smooth s-shape curve. For simplicity and numerical stability, I defined a kinematic-based primitive based on logistic-growth:
\begin{align}
    T &= \left| \text{target} - y_0 \right| \label{logistic-growth-1}\\
    S &= \text{sign}\left(\text{target} - y_0\right) \label{logistic-growth-2}\\
    A &= \frac{T}{\varepsilon} \label{logistic-growth-3}\\
    R &= \exp(r) \label{logistic-growth-4}\\
    D &= 1 + A \cdot \exp\left(-(T+R)(t-t_0)\right) \label{logistic-growth-5}\\
    y^* &= S \cdot \frac{T}{D} \label{logistic-growth-6}
\end{align}

where $y_0$ is the motor trajectory's value before the target update (at the beginning $y_0 = 0$); $\varepsilon$ is the baseline value (according to logistic growth, the initial population cannot be zero) and it is currently set to $10^{-6}$; $R$ is the positive growth rate constant (for training the parameter $r$ is left unbounded and $\exp(r)$ ensures its value is positive); $t$ is the current global time, while $t_0$ is the global time at the start of the new adjustment (the primitive is executed at the normalized time $t-t_0$). Finally, the resulting adjustment $y^*$ is the superimposed to the current motor trajectory $y$:

\begin{equation}
y = y + y^*	 \label{logistic-growth-7}
\end{equation}

There are multiple motor modules running in parallel (Fig.~\ref{fig-model-structure}), each assigned to a vehicle control (brake, throttle, and steering). These parallel modules receive the same input of size $kC$ from the preceding modules: the $kC$ spike values \{0: dormant, 1: spike\} from the accumulator module, and the $kC$ features from the perception module (Fig.~\ref{fig-model-structure}). That is, the sensory and accumulator modules create information for general motor control, without being assigned to a specific control output.

The perception features are used to compute proposals for a new motor target value via a linear function (bias and gain). The spike is used to activate (or not) the new target proposal as the proposed target value would be multiplied by either 0 or 1. There are two reasons for using the output from the perception and accumulator modules together. The first reason is conceptual. I assumed that a new target proposal would be proportional to the current perception feature's value, which would be otherwise lost across the accumulators. The second reason is practical. As the accumulator module outputs a sparse array (most often filled with 0s), training via gradient-based optimization would not perform well. By having a skip connection (inspired by He et al. \cite{heDeepResidualLearning2015}), gradient information is retained across the modules.

The proposed $kC$ targets are non-linearly compressed into a single target that will be used to execute the motor primitive. This approach resembles the inhibitory and excitatory branches described by Boda et al. \cite{bodaComputationalDriverModel2020}, but without the need to specify these pathways. Target combination is done with three linear layers with bias and mish activation in between. The parameter $r$ as well as the linear layers' parameters are learned during training, independently for each motor module.

\subsection{Data}
The model was fit on data collected on a test-track to study on how drivers negotiate an intersection with an approaching e-scooter \cite{raschRightTurnModeling2025}. The track resembled an urban intersection with a straight two-lane road and a side road for right turns. The e-scooter dummy travelled straight along a cycle lane that transitioned into an unsupervised bicycle passage crossing the side road. Twenty-five participants drove a vehicle with automatic transmission. The e-scooter dummy was controlled remotely to ensure consistency across trials. Drivers were instructed to turn right at the intersection in nine trials, each combining different speeds and e-scooter behaviors. The training runs with the stationary e-scooter were also included. Data were recorded from the vehicle and the e-scooter (e.g., position, speed, acceleration, and pedal movement) and they were synchronized at 10 Hz. All details are in the original paper \cite{raschRightTurnModeling2025}.

The following input signals to supply to the perception module were selected: 1) vehicle longitudinal speed, which is what the driver would see on the speedometer; 2) GPS position (longitude, latitude), which carries information about the position of the car on the test track and the distance to the intersection; 3) target speed for the trial, which was the speed that the drivers were asked to reach and maintain; 4) driver's gaze pitch and yaw, as recorded by the eye-tracker in the cabin; 5) e-scooter longitudinal speed; and 6) GPS position (longitude, latitude), which carries information for obstacle avoidance. In total, there were $C=9$ sensor channels to the perception module from 256 trips.

The only data processing was data standardization. The input training data were standardized with Scikit-learn's StandardScaler \cite{sklearn_api}. For the outputs, the brake and throttle output was scaled to the range $(0, 1)$, while the steering was scaled to $(-1, 1)$ using the Scikit-learn's \emph{MinMaxScaler} \cite{sklearn_api}. The scalers computed on the training data were applied to the validation set as is.

\subsection{Training}
The model was trained end-to-end with a leave-one-out driver for 750 epochs and a single batch. The optimizer was \emph{Adam} \cite{kingmaAdamMethodStochastic2017} (learning rate $3 \cdot 10^{-3}$ without scheduler). As the timeseries were of different lengths, the shorter sequences were padded. This padding was ignored (masked) when computing the loss. With the current network configuration, there were about 10 thousands parameters to train.

The model outputs three arrays (the motor trajectories for brake, throttle, and steering). The loss was calculated by, first, computing the mean absolute error (MAE) between the corresponding observations and predictions for each motor output; Second, these three errors were averaged; Finally, the average losses were summed together. This aggregated loss was used for backpropagation through time.
Training was done on the CPU (32 cores) with \emph{PyTorch-Lightning} (v. 2.5) \cite{LightningAIPytorchlightningPretrain}. The current dataset and model implementation did not benefit considerably from training on the GPU. As initial proof-of-concept, there was no comprehensive hyperparameter tuning nor (nested) cross-validation. At the end of the training, the model with the lowest validation loss was kept.

\section{Results and discussion}

\begin{figure}[tb]
    \includegraphics[scale=0.85, keepaspectratio]{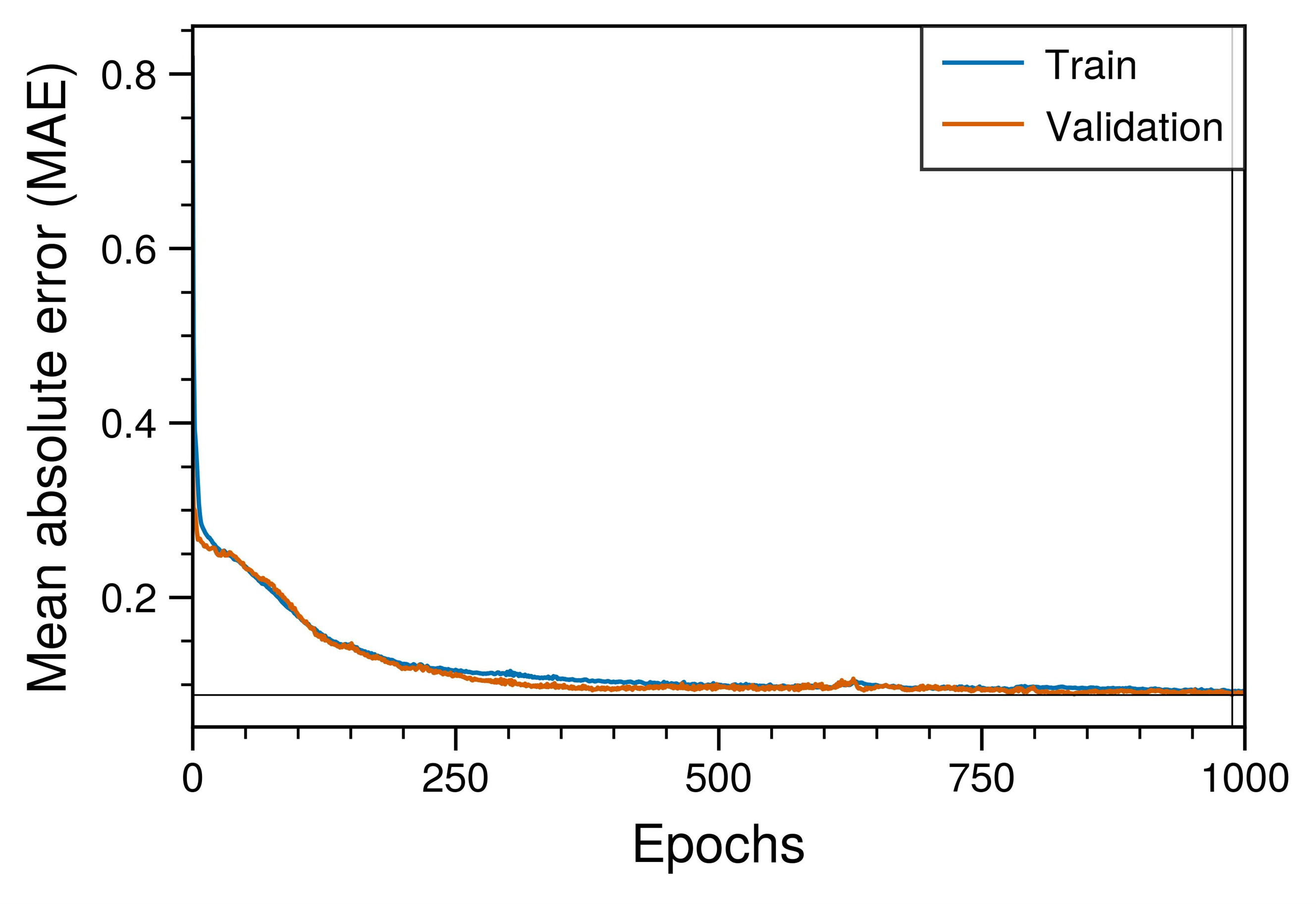}
        \centering
        \caption{Train and validation loss. The best score (min validation loss) of 0.088 was reached at epoch 988 (thin black lines).}
        \label{fig-loss-curves}
\end{figure}

\begin{figure}[tb]
    \includegraphics[width=\columnwidth, keepaspectratio]{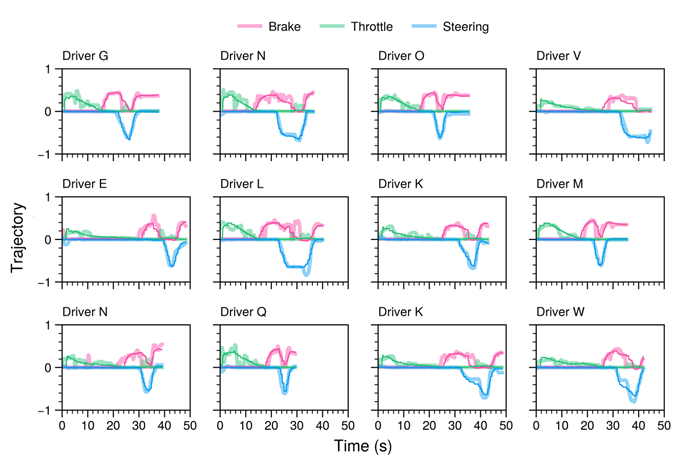}
        \centering
        \caption{A random sample from the training set (multiple drivers and multiple trials). Thicker line is the observed output; thinner line is the predicted output.}
        \label{fig-training-set-example}
\end{figure}

\begin{figure}[tb]
    \includegraphics[width=\columnwidth, keepaspectratio]{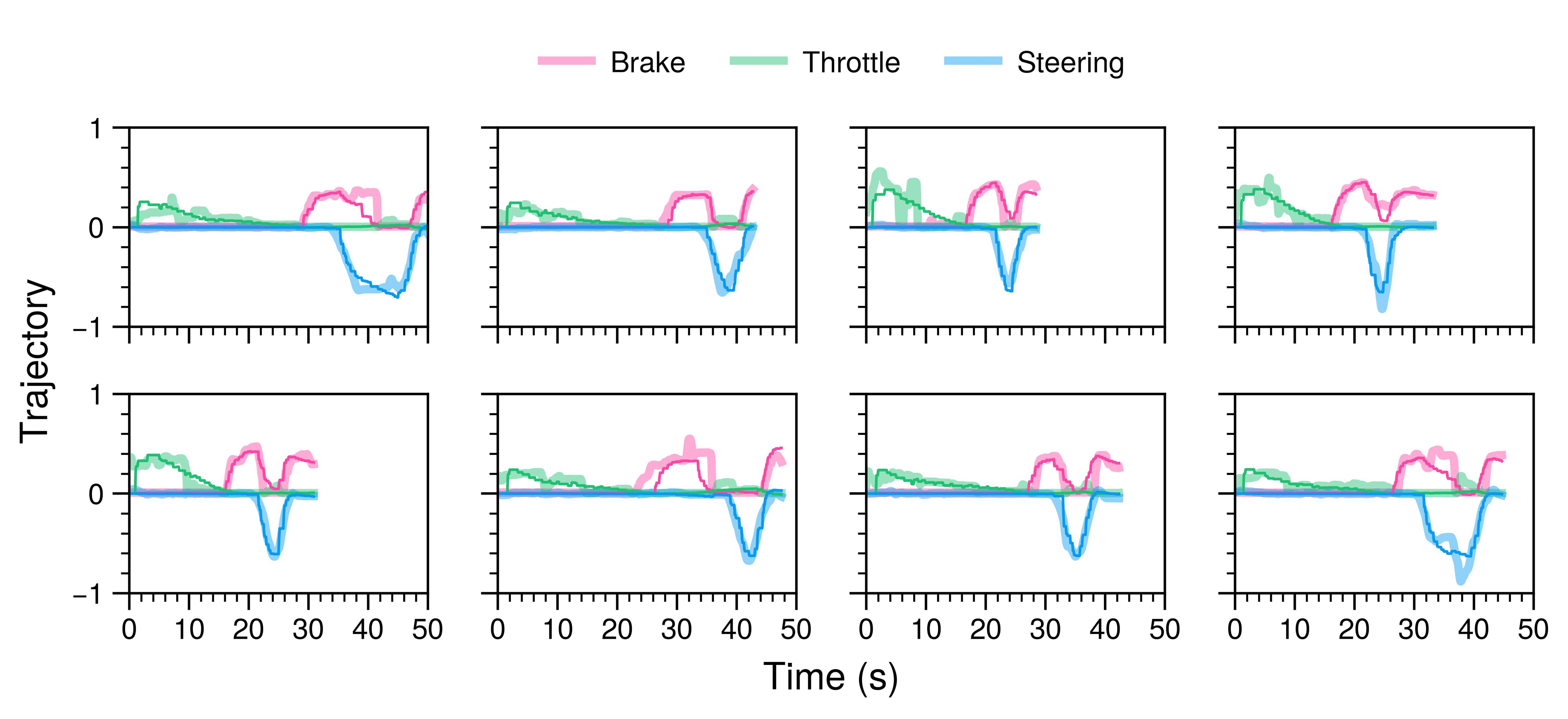}
        \centering
        \caption{A random sample from the validation set (trials from one driver). Thicker line is the observed output; thinner line is the predicted output.}
        \label{fig-validation-set-example}
\end{figure}

\begin{figure}[tb]
    \includegraphics[width=\columnwidth, keepaspectratio]{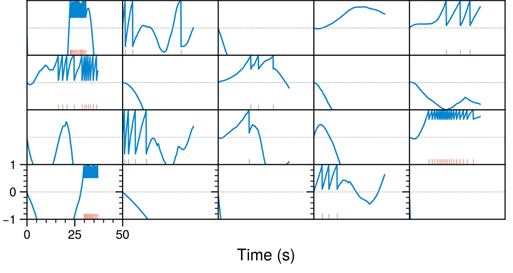}
        \centering
        \caption{Neuron potential (accumulated evidence) for a randomly selected sample in the training dataset. The red marks at the bottom of each panel are spikes.}
        \label{fig-spikes-example}
\end{figure}

\begin{figure}[tb]
    \includegraphics[scale=0.85, keepaspectratio]{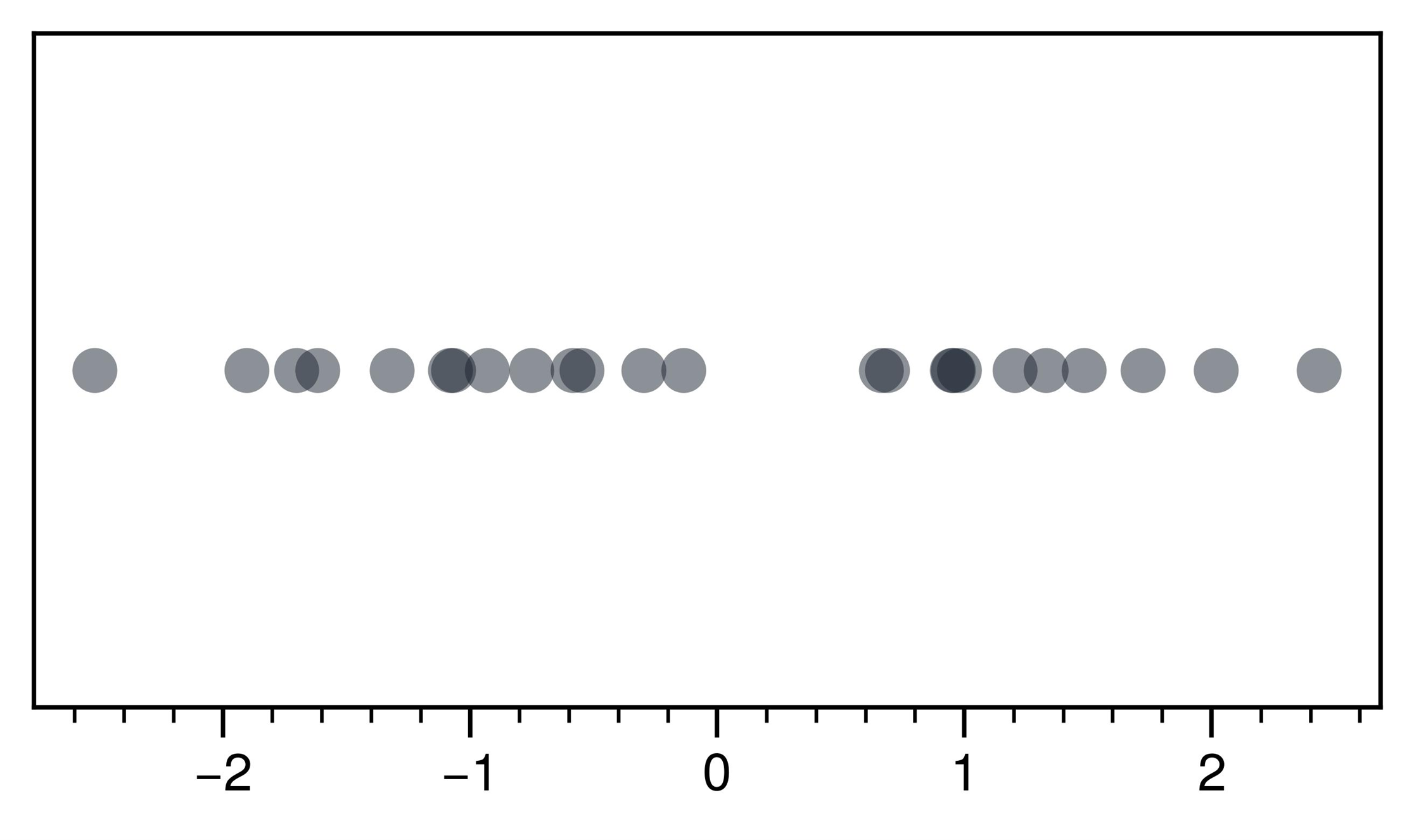}
        \centering
        \caption{A projection of the drivers embedding to a single dimension via Principal Component Analysis (PCA). Each dot is a single driver. It appears that there are two clusters (driving style).}
        \label{fig-drivers-embedding}
\end{figure}

Over the course of the training, both training and validation loss consistently decreased (best validation score was a MAE of $0.088$; Fig.~\ref{fig-loss-curves}). Overfitting on the train set was mild. Despite using a single leave-one-out driver for validation, the training and validation score remained close. This trend suggests that the data distribution between the sets was comparable, and the model generalized well.

The model accurately fitted the training data; this accuracy was maintained on the validation set across the different test-track runs—each run involved different speed limits and different interactions with the e-scooter—without being specifically designed to address these scenarios' variation. The predicted and observed trajectories for braking, throttle, and steering controls showed good temporal alignment, indicating that the model was stable over the time course of the trial. Unlike RNNs, the model does not have dedicated memory cells to retain past information. This suggests that the accumulator state was sufficient to capture relatively long-term dependencies in the data.

Generally, the model successfully captured key patterns of driver behavior such as early throttle application, sudden braking, and smooth steering. The advantage of the model over previous work \cite{markkulaSustainedSensorimotorControl2018,bodaComputationalDriverModel2020,svardQuantitativeDriverModel2017} is its ability to predict multiple controls simultaneously with little to no additional cost. While standard deep neural networks might achieve superior fit, they would typically lack the cognitively-grounded and neuroscience-informed architecture of Akkumula, which increases its explainability.

At a more detailed level, there are discrepancies in motor initiation and modulation, which warrant further investigation. For example, the model may predict braking slightly earlier (or later) than what was observed. Moreover, it failed to accurately replicate the intermittent throttle control, instead fitting an average trajectory (Fig.~\ref{fig-training-set-example} and Fig.~\ref{fig-validation-set-example}). The impact of these discrepancies on the vehicle kinematics and on the resulting interaction with the e-scooter is currently unclear. Future refinements could include a loss function that prioritizes timely motor initiation. Furthermore, a vehicle model could also be added to evaluate the error in vehicle trajectory as a result of errors in vehicle controls. These potential improvements are reserved for future work.

An advantage of models built with Akkumula is their capacity to include a large population of neurons, compared to the limited number used in earlier studies \cite{markkulaSustainedSensorimotorControl2018,bodaComputationalDriverModel2020,svardQuantitativeDriverModel2017}. Despite having many neurons, all their neuronal parameters (e.g., reset voltage, leakage) could be learned at once (instead of fixing their values for computational efficiency as often done). At this stage, the specific function of each neuron remains unclear, but future research could investigate this aspect in more detail, examining neuronal activity across different drivers and traffic scenarios (work that may also clarify the meaning of the automatically learned features).

The model allows for inspection of individual neuronal activity: which neuron spiked, when, and how often (Fig.~\ref{fig-spikes-example}). The neurons in the model exhibited diverse characteristics. Some spiked frequently at specific instants, others spiked less often but regularly during the trial, while the remaining were inactive. Furthermore, some neurons reset to a value close to 0 after firing, while other reset to a value closer the firing threshold.

The neurons in the model directly accumulated noisy signals, rather than noisy accumulation of evidence \cite{markkulaSustainedSensorimotorControl2018,svardQuantitativeDriverModel2017}. These two mechanisms may be practically equivalent. But injecting noise directly into the model could potentially impact gradient-based optimization (this aspect needs further investigation). Additionally, the neurons directly accumulated features from the perception module, instead of accumulating the prediction error (as in \cite{markkulaSustainedSensorimotorControl2018,bodaComputationalDriverModel2020}). Defining prediction errors is not trivial when dealing with multiple, automatically extracted perceptual features—attempting to do so would also likely involve a series of assumptions that could reduce the adaptability of the model. While error-driven control adjustments may increase the ecological validity of the mode even further (see free energy principle \cite{fristonFreeenergyPrincipleRough2009}), at this stage the model works well in practice without it. It is also possible that the perception module inherently created signals signal resembling a prediction error. Nevertheless, future work could implement such error-driven mechanism, perhaps by devising a module similar to the Residual Blocks in Residual Networks \cite{heDeepResidualLearning2015}.

The personalization module reveals a separation of drivers into two clusters (Fig.~\ref{fig-drivers-embedding}). Analyzing the same dataset, Rasch et al. \cite{raschRightTurnModeling2025} observed that some drivers had a more aggressive behavior compared to others, and they did not yield to the approaching e-scooter. This clustering pattern could represent a fundamental distinction between cautious and aggressive driving styles \cite{sagbergReviewResearchDriving2015}. This interpretation is preliminary; future research could specifically investigate whether these embeddings reliably capture and differentiate distinct driving behaviors using a larger population of drivers with known driving styles (e.g., risk taking propensity).

\section{Conclusions}
Akkumula enables the creation of driver models that are both ecologically plausible and computationally efficient. The core idea behind the framework is that driving is a continuous perception-action process, where evidence from the environment builds up to initiate motor actions \cite{markkulaSustainedSensorimotorControl2018}. The novel aspect of Akkumula is in bridging knowledge across research domains, continuing the interdisciplinary approach from previous work on the topic \cite{markkulaSustainedSensorimotorControl2018,bodaComputationalDriverModel2020}. Markkula et al. \cite{markkulaSustainedSensorimotorControl2018} and Boda et al. \cite{bodaComputationalDriverModel2020} integrated formerly separate concepts—neuronal evidence accumulation, excitatory/inhibitory stimuli, intermittent control, and motor primitive superposition. This paper extends that line of work and implements it within deep learning architectures while maintaining a degree of explainability, a property often lost in deep neural networks. This means that models built with Akkumula could be used for understanding the underlying behavioral mechanisms and not just for prediction. The framework offers greater adaptability, modularity, and efficiency compared to existing solutions. The initial results are promising and there are many opportunities for improvement to make accurate driver models that can be used for virtual safety assessment.

\emph{Adaptivity}. Akkumula does not require extensive data preprocessing—except for data normalization to improve training efficiency. Unlike previous work, it is not necessary to pre-define the perceptual features for accumulation (e.g., looming \cite{markkulaSustainedSensorimotorControl2018,bodaComputationalDriverModel2020,svardQuantitativeDriverModel2017,bianchipiccininiHowDriversRespond2019}) or to identify the control adjustments in the output motor signal (as in \cite{markkulaSustainedSensorimotorControl2018,bodaComputationalDriverModel2020}). The model can accommodate additional (or fewer) inputs/output signals, including different data types (e.g., images) with only minor changes. This flexibility offers the advantage of minimizing arbitrary design choices and making the model applicable across various scenarios.

\emph{Modularity}. Akkumula is written using PyTorch \cite{anselPyTorch2Faster2024}, an established programming framework, and SpikinJelly \cite{fangSpikingJellyOpensourceMachine2023,fangwei123456Fangwei123456Spikingjelly2024}, an open-source library. Both PyTorch and SpikinJelly provide many building blocks for designing complex neural networks. The modules in Akkumula are designed for easy organization and replication into different architectures. This has the advantage of eliminating the need for specifying exact pathways of evidence accumulation and motor control (as in \cite{markkulaSustainedSensorimotorControl2018,bodaComputationalDriverModel2020}). Akkumula can be combined with other types of ANNs, such as Convolutional Neural Networks to process data coming from cameras, or other types of signals. These capabilities make Akkumula easy to integrate in larger (or pre-existing) networks.

\emph{Efficiency}. Akkumula enables end-to-end training using batches of data. This means that the raw initial inputs and final outputs from the entire dataset are supplied to the model as is, and its parameters are optimized to directly minimize the difference between the observed and predicted motor outputs. Gradient-based minimization of errors through gradient descent is efficient and can optimize models with a vast number of parameters. This capability allows Akkumula to handle large datasets and complex architectures. Moreover, thanks to PyTorch's native-support, training can be accelerated using GPUs, which can be particularly beneficial when processing images and videos.

\section*{Acknowledgments}
I thank my colleagues at Autoliv Research for their comments and suggestions on this work. The work started within the project e-SAFER (grant number 2022-01641) financed by FFI (Sweden's Strategic Vehicle Research and Innovation program) and then continued with fundings from Autoliv.

\bibliography{IEEEabrv,main.bib}

\end{document}